\documentclass[conference]{IEEEtran}
\IEEEoverridecommandlockouts
\usepackage{cite}
\usepackage{url}
\usepackage{amsmath,amssymb,amsfonts}
\usepackage{algorithmic}
\usepackage{graphicx}
\usepackage{textcomp}
\usepackage{xcolor}
\def\BibTeX{{\rm B\kern-.05em{\sc i\kern-.025em b}\kern-.08em
    T\kern-.1667em\lower.7ex\hbox{E}\kern-.125emX}}
\begin{document}

\title{Indoor Localization for Autonomous Robot Navigation}

\author{\IEEEauthorblockN{Sean Kouma}
\IEEEauthorblockA{\textit{Computer Science} \\
\textit{Colorado State University}\\
Fort Collins, Colorado \\
skouma@colostate.edu}
\and
\IEEEauthorblockN{Rachel Masters}
\IEEEauthorblockA{\textit{Computer Science} \\
\textit{Colorado State University}\\
Fort Collins, Colorado \\
ramast1@colostate.edu}
}

\maketitle

\begin{abstract}
Indoor positioning systems (IPS’s) have gained attention as outdoor navigation becomes prevalent in everyday life. Research is being actively conducted on how indoor smartphone navigation can be accomplished and improved using received signal strength indication (RSSI) and machine learning (ML). IPS’s have more use cases that need further exploration, and we aim to explore using IPS’s for the indoor navigation of an autonomous robot. We collected a dataset and trained models to test on a robot. We also developed an A* path-planning algorithm so that our robot could navigate itself using predicted directions. After testing different network structures, our robot was able to successfully navigate corners around 50\% of the time. The findings of this paper indicate that using IPS's for autonomous robots is a promising area of future research.
\end{abstract}


\section{Introduction}
Improvement and applications of indoor positioning systems (IPS’s) is an active area of research increasingly involving machine learning (ML). Specifically, there is an interest in using IPS’s for indoor navigation applications on smartphones, but the technology has been lacking compared to outdoor smartphone navigation systems like Google Maps. Current research is improving indoor smartphone navigation, but there are other interesting use cases for IPS’s that need further research to understand and implement effectively. Our research examines one of those use cases, which is to use IPS’s to navigate an autonomous robot. 

Determining outdoor position is relatively easy as GPS can be used. Unfortunately, GPS does not work well for calculating indoor position due to poor satellite signal indoors\cite{b5}, especially when building layouts are complex and detailed with multiple floors and wings \cite{b2}. Wireless fingerprinting using received signal strength indicators (RSSI) has become a popular method of calculating indoor position, making use of routers within a building to get relative positions to all routers for the calculation of overall position. However, RSSI alone has its limitations. Primarily, noise can interfere with accuracy, especially when there are few access points to retrieve RSSI data from \cite{b1}. Fortunately, ML techniques can help mitigate noise while learning useful information for accurate location prediction. For example, state of the art frameworks using DNNs and different CNNs that are optimized for mobile have shown success for smartphone navigation \cite{b1,b2,b3,b4}. These frameworks account for size, energy consumption, and other optimizations essential for mobile deployment. 

Existing frameworks are primarily used in the context of indoor navigation on smartphone rather than autonomous navigation using indoor robots. Autonomous navigation presents another set of challenges when developing an effective ML model. Since the robot is on the ground rather than being held by a human, the way that it receives RSSI data may differ slightly. For example, depending on the way the walls of a certain building may be structured, an object lower to the ground may experience more interference. Additionally, more error is tolerable when a human is navigating, as humans can recognize when directions are slightly incorrect and use reason to fill in where technology falls short. For an autonomous robot, accuracy is critical as the robot will drive based on the prediction, whether or not that prediction is correct. Thus, poor models may cause the robot to crash and potentially damage expensive equipment. Using video feeds to predict steering angles is one possible approach to autonomous driving. Autonomous navigation with a location-aware device is a more complicated task that will require the use of RSSI. 

In order to solve the problem of indoor autonomous robot navigation, our overall approach starts with an implementation based on existing work, which we will then modify to improve the accuracy of our solution. Data needs collected in the building of interest, then fed through a basic network, then tested on the robot. The network then needs modified with prior work in mind to find the best solution for this use case. Through thoroughly documented trial and error, we will converge on the necessary components for an optimal solution, then ensure that the performance is replicable.

\subsection{Problem Statement}
While extensive work has been done regarding both indoor localization and autonomous navigation, there is an intersection between these two fields we hoped to address. Some of the recent advances in indoor localization are RSSI based approaches and we wanted to see how well this approach would extend to an autonomous navigation system. Our goal for this project was to use a Deep Neural Network (DNN) on an embedded system to solve the problem of autonomous, indoor navigation using robots. Along with that, we wanted our solution to meet some additional criteria as well. First, we did not want our robot platform to use any unusual or non-standard components that are not available on most embedded devices today. Secondly, we wanted our process and solution to be extendable to any location or building with multiple access points that are broadcasting a Wi-Fi signal. It is expected that the more access points that are available, the more accurate our predictions will be, but there is not a specific minimum number of access points required. Finally, we wanted our solution to not be computationally intensive, so that it is feasible to deploy on embedded hardware.

\section{Prior Work}
With the success of outdoor localization and navigation using GPS, there has been increasing interest in developing a similar method of navigating complex indoor environments. Unfortunately, GPS is challenging to use indoors due to complex environments blocking signals. GPS cannot achieve the level of granularity needed to successfully navigate in buildings, especially those with multiple floors, so the prior literature has focused on developing new techniques for determining location indoors, then applying those techniques to create navigation frameworks.
\subsection{Indoor Localization Using Wi-Fi}
Wi-Fi-based indoor localization has been a popularly researched technique because Wi-Fi is standard in many buildings, and Wi-Fi access points (AP) provide signals that can determine location with respect to an AP. Namely, RSSI values collected through Wi-Fi fingerprinting give indications of proximity to AP’s, which can be treated as distance values and mapped to a grid. Prior work has focused on processing and interpreting RSSI values to create machine learning frameworks that can estimate position indoors with the best accuracy possible. 

Accuracy is a key challenge with Wi-Fi fingerprinting because walls, furniture, and changes in object placement can all cause signal variations that negatively affect accuracy. An analysis performed on Wi-Fi fingerprinting techniques found that information provided by weak access points can help improve accuracy, that varying the rotation of a device during signal collection may help improve accuracy but did not in their experiment, and that using both 2.4 and 5 GHz bands improves accuracy\cite{b9}. While these considerations can help make ML models more robust, research at the framework level has shown potential for models that are more robust by design and optimal for localization. Guo et al created a modified version of Wi-Fi fingerprinting where they used a global fusion profile to fuse together a group of fingerprints. \cite{b10} Typical Wi-Fi fingerprinting is susceptible to environment changes affecting accuracy, so they applied information fusion, which helped with single fingerprint localization in the past, to a multi-fingerprint localization approach. This fusion technique used RSSI as well as signal strength difference (SSD) and hyperbolic location fingerprint (HLF) to create a system robust to RSSI fluctuations.

Even more recently, indoor localization using Wi-Fi techniques has seen great improvement through the development of full frameworks and expediting of data collection. The DLoc algorithm was created to be used with a mapping platform called MapFind, which allowed smartphones to access the map of an indoor environment and determine their position relative to the map\cite{b11}. MapFind was made using a robot equipped with LIDAR and an odometer. MapFind was created as a data-driven approach to indoor localization that automatically generated data to train DLoc. Additionally, DLoc was modeled with one encoder and two decoders, one decoder maintaining access point consistency while the other estimated location. This research created an end-to-end framework for smartphone localization indoors, however, there were a few limitations like speed of data collection and scalability testing to improve in the future. In the past year, Yin and Lin created a ML based IPS that used a mathematical formula to create Wi-Fi fingerprinting data\cite{b5}. Since collecting fingerprints for large spaces is time consuming and impossible in some instances, Yin and Lin used four anchor points as references and created mathematical formulas for four models to measure the distance between the anchor points and some test location point in different ways. Then, they trained and tested on the predicted and real distances between the test location and anchor points, eliminating a lot of the tedious measurements traditionally taken.

Despite new developments in the field, one challenge that advancements in indoor localization struggle to overcome is the overhead of embedded deployment. Often, more accurate Wi-Fi techniques struggle to achieve high accuracy without being computationally expensive. Additionally, there has not yet been a fingerprinting-only solution that has achieved good accuracy while still being deployable on embedded devices. The CNN-Loc framework was created to use Wi-Fi fingerprints as inputs to a CNN model\cite{b1}. CNN-Loc used greyscale images created from RSSI values, which were created using Hadamard Products. CNN-Loc was also created using a hierarchical classifier, which made the framework scalable. After the creation of this CNN-based framework, research continued on how to maintain accuracy while deploying on mobile devices. The CHISEL framework was proposed to keep the accuracy and robustness of deep learning solutions using Wi-Fi fingerprinting while also making those solutions deployable on embedded devices\cite{b4}. Convolutional neural networks (CNNs), show great promise for handling RSSI data, yet they are computationally complex and consume memory at infeasible rates for embedded use. CHISEL used state of the art deep learning techniques combined with state-of-the-art model compression techniques to preserve model accuracy while optimizing models for mobile usage.
\subsection{Indoor Localization Using Other Techniques}
Bluetooth fingerprinting is another technique similar to Wi-Fi fingerprinting. Bluetooth low energy (BLE) is a popular method because of its affordability and how easy it is to deploy BLE beacons. Sthapit et al used a BLE machine learning approach, creating a radio map of the navigable area and collecting RSSI values from the sensors in that area in an offline phase, collecting data and training a model during an online phase, then doing real time testing of the system\cite{b14}. They were able to estimate location with an average error of 50cm, but their testing was limited to a corridor and their lab room. Koutris et al recently used BLE for indoor localization and made improvements on existing systems, using raw in-phase and quadrature-phase (IQ) values in addition to RSSI values to better estimate angle of arrival (AoA) via multiple anchors\cite{b15}. Additionally, they tried a variety of different neural network architectures to determine which model generalizes best, treating anchor points independently, fully jointly, in tuples, and jointly as measured by a CNN. Overall, BLE is still a popular topic of research for indoor localization, it is just more susceptible to variations in signal than Wi-Fi fingerprinting, which presents an additional challenge for maintaining accuracy\cite{b14}. There are other methods for indoor localization, like radio signal propagation to detect proximity to a Wi-Fi point, but developing these models is complex and tends to lead to poor accuracy\cite{b12,b13}.
\subsection{Indoor Autonomous Robot Navigation Systems}
Indoor autonomous robots are an application of indoor localization technology. Localization for autonomous, indoor robots was first popularized in 1988. At AT\&T Bell Laboratories, Cox developed a position system for robots called Blanche, which used odometry and matching to map the range data onto an existing 2D map of the environment to correct error\cite{b16}. This work was groundbreaking because it was an efficient, low-cost way of doing indoor localization in a fixed space with a known map. After this work, there was an interest in using robots to help humans perform daily tasks with more ease. In 2005, the researcher Hayashi worked toward creating an autonomous robot that could do practical tasks in a fixed environment in order to aide humans in daily activities\cite{b17}. Hayashi developed a robot that would navigate using a knowledge base and sensors, tackling issues of processing contextual information, and using that information to make decisions. Hayashi found that dead-reckoning was not a good method for indoor robots that are intended to be aides to humans because the robots have move safely with humans without crashing\cite{b18}. Hayashi proposed that a robot must have object detection and self-localization in order to accomplish safe navigation, creating a navigation system using only an ocellus camera with no localization techniques like Wi-Fi fingerprinting. Hayashi’s robot was able to successfully navigate rooms and corridors while recognizing and correcting distance errors. It is unclear if the robot could comprehend advanced, multiroom directions in complex buildings, as the application is in a single, simple room or corridor. 

For robots that are designed to travel across multiple rooms, Wi-Fi fingerprinting can help understand complex building layouts. Recently, Khanh et al proposed a new framework for autonomous robots, arguing that rather than using Wi-Fi fingerprinting across many access points, using a cloud navigation system based on a single Wi-Fi access point can determine position more efficiently and accurately\cite{b14}. Efficient and accurate positioning can help some with the operational safety concerns Hayashi presented, but improvements in object detection and collision avoidance are also being made for safer navigation. Recently, zero-shot object detection (ZSD) was proposed as a solution\cite{b19}. Essentially, object detection is a challenging topic indoors because of the number and complexity of object classes that exist in buildings. Since it is unrealistic to train a network for every object that can possibly be in a building, Abdalwhab and Liu introduce ZSD for robots to train a network on some classes then see how well the robot predicts classes that it has never seen before. Combining object detection and self-localization techniques is a promising option for robot navigation but also increase the complexity of the solution, which still needs to be optimized for deployment on mobile devices. This issue is still a subject of current work.

\section{Data}
\label{section:data}
We are working with RSSI data for our project. RSSI is an indicator of how strongly a device receives a signal from an access point within range. (In our case, we used Wi-Fi access points as opposed to Bluetooth). During our project we collected our own RSSI data for our project and we collected our data from the fourth floor of the CSU CS building. Specifically, we used the Bash command "iwlist wlan0 scan" to collect the RSSI values from the access points in the building. Note that we only collected data for the SSID's 'CSU Net' and 'CSU Visitor' to ensure that we were not picking up any mobile hotspots or other unstable networks. Additionally, due to the level of precision needed to successfully drive our robot autonomously, we made the decision to take a new measurement every square foot. We randomly picked a section of the floor which included a corner and took 95 square feet worth of data. To avoid one-off errors, we re-sampled the RSSI values three times at each location to ensure that we received a reliable measurement.

After collecting the data, we did a few things to preprocess and filter the data from the raw values. We developed scripts to compile the values from all the individual files that we generated from running “iwlist wlan0 scan” and we put our processed data into a CSV file. The CSV file represented a single two-dimensional matrix where each row was the RSSI values collected at a certain location. The columns were the MAC addresses of the access points within range except for the last two columns which were the x and y coordinates of the location where the data was collected (which we mapped and recorded as we collected the RSSI values). Finally, if our robot was only able to sense some access points in some locations, then we replaced all of the missing values for the missing access points with zeros. Once we finished these preprocessing and filtering steps, our data was ready to be fed into our network.

\section{Methods}
\begin{figure}[htbp]
\centerline{\includegraphics[scale=0.5]{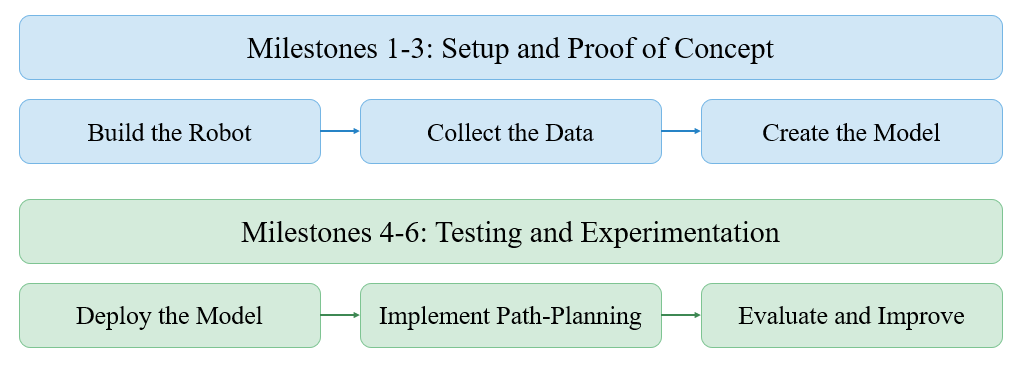}}
\caption{Visualization of Project Process}
\label{methodsfig}
\end{figure}

To build a solution that meets the criteria outlined above, we broke our problem up into separate individual milestones. 
\begin{enumerate}
    \item Find and build the robot platform to collect data and demonstrate our results.
    \item Figure out how to collect RSSI data on our robot platform.
    \item Develop a machine learning model to predict the current position of the robot based off of RSSI signals. 
    \item Deploy this model in a real-time test environment.
    \item Develop a path-planning algorithm which uses the predicted location data to navigate to a desired destination.
    \item Evaluate our results.
\end{enumerate}
Figure~\ref{methodsfig} shows a visualization of our process.

\subsection{Building the Robot Platform} 
\begin{figure}[htbp]
\centerline{\includegraphics[scale=0.1]{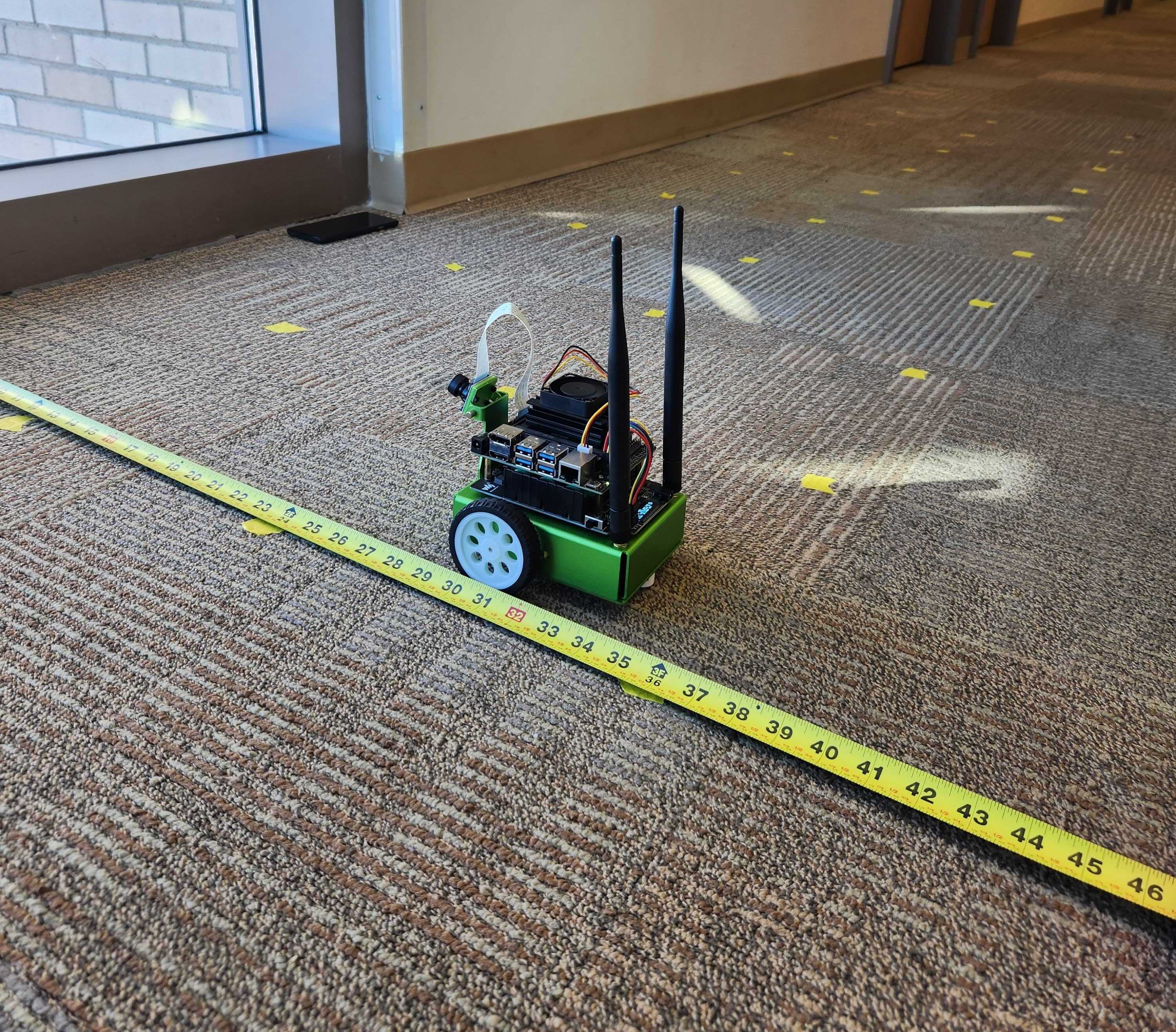}}
\caption{Our Robot}
\label{fig}
\end{figure}

Because we wanted our model to be able to work in real-time, we wanted a robot built off of the Nvidia Jetson Nano since it performs better in ML tasks \cite{b6}. This narrowed down our options significantly, but ultimately we chose the Waveshare Jetbot AI Kit for our project. This kit had everything we needed including a mobile base, WiFi antennas, a remote allowing you to manually drive around the robot, and finally it included a camera in case we wanted to expand on our Wi-Fi Fingerprinting approach to incorporate object detection in future work outside the scope of the class\cite{b7}. Building the robot and getting it functional took several hours, and we did not realize in advance that the kit did not include the unusually-sized rechargeable batteries needed to power the platform. Ultimately, we got the robot built as shown in Figure~\ref{fig} and were able to remotely run Bash and Python commands through Jupyter Notebook using a remote interface. Getting to this point allowed us to proceed to the next step.

\subsection{Collecting RSSI Data} Collecting the RSSI data from our robot proved to be more involved than we anticipated. First of all, a substantial amount of time was spent searching for an adequate Python package that would easily provide the RSSI values along with the corresponding MAC addresses and SSID's of the networks in range. We settled on using the \textit{rssi} package \cite{b8}. Unfortunately, this package ended up not working very well with our robot. Some unknown configuration issues caused the output to be formatted strangely and included a number of strange characters and symbols. Instead, we ended up using the Bash command "iwlist wlan0 scan" which provided extensive information about the access points within range. We found that the \textit{rssi} package may have actually just been a wrapper for this Bash command.

After figuring out a method to measure the RSSI signals, we then chose the fourth floor of the Colorado State University Computer Science building as our test location due to its fairly simply layout. The floor is one big square which only has hallways running along the outside. We then collected the data as described in Section~\ref{section:data}.
\subsection{Developing a Machine Learning Model} We finally reached the point in our project where we were able to actually develop a machine learning model to use the data that we had collected. In order to develop a model that would work well, we did some experimentation. At first, we tried dropping all columns in our data which contained less than 75\% of entries. This was to ensure that our model was not trained on any bogus data that likely would not appear when we deployed our robot. Our model's accuracy level using only the columns with more than 75\% of entries was not what we had hoped. We ended up going back and instead computing a Pearson Correlation Coefficient (PCC) between all the columns and keeping those which had a positive or negative correlation of at least 0.24 with the x or y column. This resulted in only 6 remaining columns in our data. From this data, we split our data so that 75\% of the data was train data and the remaining 25\% was test data.

\begin{figure}[htbp]
\centerline{\includegraphics[scale=0.65]{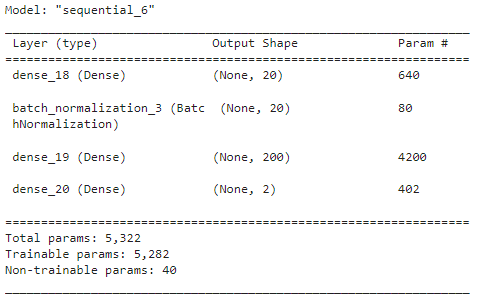}}
\caption{DNN Model Summary}
\label{fig1}
\end{figure}

While some current approaches to the indoor localization problem involve using CNN's \cite{b1}, we decided to start with a DNN to see how well it performed. Provided with more time, we would like to go back and try a CNN-based approach where a 2D image is constructed through a Hadamard product as described here \cite{b1}. We did not start with the CNN approach because we had seen good results from using DNNs, and we did not have the required prior experience with Hadamard products to implement the solution within the given time frame. We experimented with quite a few different network structures on our data. We started with a simple, one layer network that had 20 neurons, as a proof of concept to feed our data through. After getting that to work, we experimented with adding layers, and we found that adding more than three dense layers did not help improve accuracy. We experimented with increasing and decreasing the number of nodes in the layers, and we found that having a larger number of nodes in later layers helped with learning. Through trial and error, we ended up with a model like what's shown in Figure~\ref{fig1}. As is visible, we used three dense layers with the last layer predicting the x and y coordinates of the robot. We experimented with batch normalization and found that it helped if placed between the first and second layer.

This model was trained for 700 epochs using a Mean Average Error (MAE) loss function. Additionally, while fitting the model we set the validation split to be 0.2 so that we could get an idea of how well our model would perform on our test data.

While originally we had planned on using quantization, compression, and pruning techniques to reduce the size of our model and improve inference time, they were not very helpful for this particular model. In practice, we had plenty of space on the Jetson Nano to load our model into memory and the amount of time it took to make an inference was negligible compared to measuring the RSSI values. Keeping pruning and quantization would have decreased the accuracy of our model and impacted the operational safety of our robot, which was not worth it for this use case. Thus, we made the decision to spend our time on other parts of the project rather than perfecting model compression. However, in the future, when a more complex model is developed to achieve even greater accuracy, pruning and quantization will likely be needed in that situation, this model just happened to be relatively small and susceptible to changes in accuracy from pruning and quantization.

\subsection{Deploying the Model in a Real-Time Test Environment}
To deploy the model in a real-time environment, we saved the trained model using Keras and loaded the model onto the Jetson Nano. While attempting to load the model, we ran into numerous unexpected issues with conflicting versions of Tensorflow and different environment issues. The Nvidia Jetson Nano has a special version of Tensorflow that has been designed to perform optimally on the Jetson's limited processing power, but these version differences caused significant issues with loading the saved Keras model. Additionally, there also was a bug where the OS image for the Jetson did not use the full space of the 256GB SD card that we were using, which meant that we needed to reinstall the entire operating system and all Keras' dependencies which took some time. Ultimately, we were able to load the saved version of the model and make inferences on the Jetson. The time it took the Jetson to do a forward pass through the network was actually negligible, but running the 'iwlist wlan0 scan' command actually took some fairly significant time (up to 2 seconds). This was rather unfortunate since the goal of the project was to continually make new predictions, but we got around this issue by making the robot wait to move until it receives a new location prediction. For example, when the robot computes a prediction of its current location, it moves forward 2 feet and then stops and waits until another prediction is generated before moving on. While this is by no means ideal, we did not have the time to try and research how to speed up the network scans as it was not the primary goal of the project.

In addition to the issue regarding slow location predictions due to the length of time it takes to do a network scan, we faced another larger issue. As is discussed in the Results section of this paper, the robot was not able to make location predictions with the accuracy we had hoped for. Originally, our intent was for the robot to continually make new location predictions and to tweak its current heading based off of each new prediction, but we realized that this was not going to be feasible. After experimenting to find another solution, we decided to switch to an approach that used checkpoints. With the new approach, a series of checkpoints are generated from the initial location leading to the goal location. The robot assumes that it is already directly facing forwards and the robot continues forwards up until it receives a prediction that it has reached its next checkpoint. Once the checkpoint is reached, it will take an action associated with that checkpoint, such as turning 90 degrees to go down a new hallway. This is not ideal, but because our accuracy was not as good as we had hoped it seemed necessary.
\subsection{Developing a path-planning algorithm to navigate to a desired destination}
At this point in the process, our robot had the ability to navigate to checkpoints, but we wanted to be able to give it a destination location where it could map out a path to its goal and then navigate there. While a variety of path planning algorithms exist that accomplish this, we chose one of the simplest most efficient algorithms, namely the A* algorithm. Simply put, the A* algorithm needs a grid or graph as an input, an initial and goal location, and then it finds the optimum path from the initial location to the goal location. The way it finds the optimal path is through a heuristic provided by the developer which evaluates whether a step in a certain direction is closer to the goal than the current location. Not only does it look at whether a move brings it closer to the goal, but it also gives it a quantitative value representing exactly how much closer it should bring it to the goal. In our case, because we are working with a grid we can use the Manhattan distance metric for our heuristic. The algorithm then quantitatively compares its options and picks the best one (the highest or lowest value depending on how the heuristic has been built). We found a grid-based, Python approach available on Medium and successfully generated some checkpoints for the robot to navigate to \cite{b21}.

\subsection{Final Testing}
After everything was put together, it was time to put it all to the test through some final experiments. We returned to the fourth floor of the CSU CS building where we originally collected the RSSI values at various locations. We set a destination for our robot to navigate to and we ran the code. Unfortunately, we discovered an issue which had missed detection or had not existed earlier in our research, namely that when our robot was supposed to go straight, one of the wheels would rotate faster than the other one causing the robot to slowly veer to the left. This required a decent amount of calibration, but we were able to get it fixed (or rather, we would tell our robot to turn slightly to the right from time to time to offset it). A second issue we ran into at this step, was that we needed to be able to tell our robot to make 90 degree turns to be able to go down new hallways. The library we used to control the robot only allowed us to set the speed of the individual wheels and then tell them to start and stop with a separate command. Thus, we had to figure out the proper amount of time to turn one of the wheels at just the right speed to perform a 90 degree turn.

After the aforementioned issues were fixed, we finally had a robot with everything required to navigate a building. We ran through many trials and the robot was able to successfully navigate corners in the building roughly 50\% of the time.

\section{Results}
After going through the process above, we ended up with a number of different results. A tangible, physical result of our project is that we have a working robot that is able to be controlled by either a physical controller or a Jupyter Notebook executing movement commands. This robot is very flexible and can be used for research in the future and has a decent amount of computing power with the Nvidia Jetson Nano. Another result of this project is that we have the RSSI data that was collected on the fourth floor of the CSU CS building. This data can be used in future networking related projects and contains a lot of useful information. Our machine learning model development phase also yielded an all-encompassing Jupyter Notebook that can be used to test and save models on our RSSI data. This can be reused in the future for further model development.

\begin{figure}[htbp]
\centerline{\includegraphics[scale=0.48]{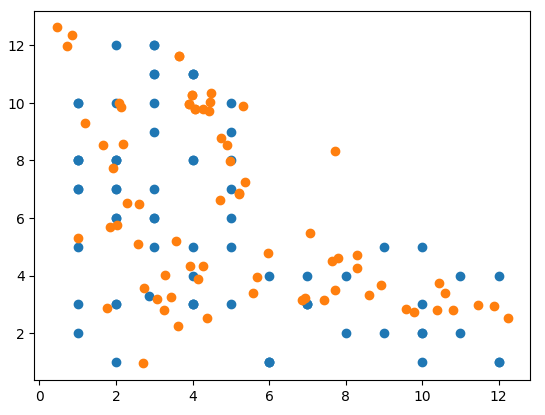}}
\caption{Predicted x--y coordinates vs actual}
\label{fig2}
\end{figure}

While the results above are meaningful, they primarily highlight the data and tools that we had to generate for this project. We have a number of quantitative results as well. First and foremost, our machine learning model resulted in a normalized Mean Absolute Error (MAE) of 0.14, which roughly equates to the robot's mean location prediction being 1.68 feet away from its actual location.  This was not the level of accuracy we had originally hoped for and was the reason why our robot was only able to successfully navigate corners 50\% of the time, but it still demonstrated that our model learned some sort of correlation.  Shown below in Figure~\ref{fig2} is a visual representation of our model's performance on our test data. The blue dots represent the actual location of the robot when the RSSI values were measured while the orange dots represent the predicted locations.

\begin{figure}[htbp]
\centerline{\includegraphics[scale=0.48]{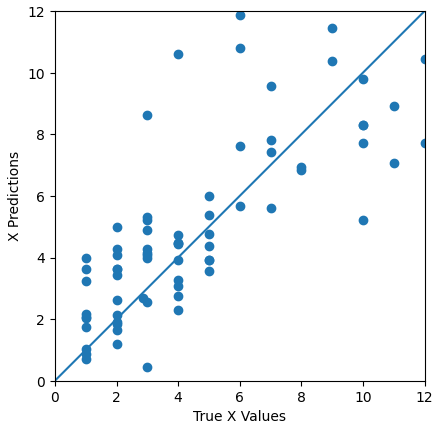}}
\caption{Predicted x coordinates vs actual}
\label{fig3}
\end{figure}

\begin{figure}[htbp]
\centerline{\includegraphics[scale=0.48]{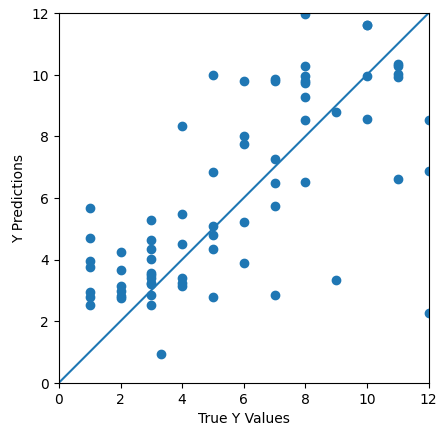}}
\caption{Predicted y coordinates vs actual}
\label{fig4}
\end{figure}

While our model did not learn as much as we had originally hoped, there is still a close enough relationship that it was usable. Additionally, while Figure~\ref{fig2} demonstrates the range and shape of our overall predictions, Figure~\ref{fig3} and Figure~\ref{fig4} below do a better job of displaying how close our predictions are to the actual x and y coordinates we defined. As mentioned above, a further research opportunity would be to see how a CNN is able to perform on our data. \cite{b1}.

\section{Conclusion}
Looking back at our research, we present both the tools that we built and used along with our quantitative results showing that there is likely a future for autonomous navigation based off of measuring RSSI values and feeding them into a ML model. Use cases for our research could include a document delivery robot inside of an office setting, an automated waiter that delivers your food to your table as soon as it is ready, or an automated security robot able to patrol the inside of buildings and record any suspicious behaviour. We believe that all of these use cases are entirely feasible and may be implemented in the future after more research is done. While we believe that further research in this area should be done, primarily regarding building a model with better accuracy and building a device-agnostic model \cite{b22}, we are optimistic about the research that we have accomplished and are hopeful that other researchers will continue to build on the work that has been done in this field.


\begin{thebibliography}{00}
\bibitem{b1} A. Mittal, S. Tiku, and S. Pasricha, ``Adapting convolutional neural networks for indoor localization with smart mobile devices,'' Proceedings of the 2018 on Great Lakes Symposium on VLSI, 2018.
\bibitem{b2} A. Nessa, B. Adhikari, F. Hussain, and X. Fernando, ``A survey of machine learning for indoor positioning,'' IEEE access. vol. 8, pp. 214945--214965, 2020.
\bibitem{b3} V. Ugave, ``Smart Indoor Localization Using Machine Learning Techniques,'' Thesis, not published.
\bibitem{b4} L. Wang, S. Tiku, and S. Pasricha, ``CHISEL: Compression-Aware High-Accuracy Embedded Indoor Localization with Deep Learning,'' IEEE Embedded Systems Letters. vol. 14.1, pp. 23--26, 2021.
\bibitem{b5} A. Yin and Z. Lin, ``Machine Learning aided Precise Indoor Positioning,'' arXiv preprint arXiv:2204.03990, 2022.
\bibitem{b6} S. Pasricha, ``Embedded Systems and Machine Learning Introduction and Logistics'' Colorado State University CS 528 Course Slides, not published.
\bibitem{b7} Waveshare ``Jetbot AI Kit'' Available: \url{https://www.waveshare.com/wiki/JetBot_AI_Kit}. Accessed October 2022.
\bibitem{b8} PyPI ``RSSI Python module'' Available: \url{https://pypi.org/project/rssi/}. Accessed October 2022.
\bibitem{b9} G. Jekabsons, V. Kairish, and V. Zuravlyov, ``An Analysis of Wi-Fi Based Indoor Positioning Accuracy,'' Computer Science vol. 47, pp. 1407-7493, 2011.
\bibitem{b10} X. Guo, L. Li, N. Ansari, and B. Liao, ``Accurate WiFi localization by fusing a group of fingerprints via a global fusion profile,'' IEEE Transactions on Vehicular Technology, vol. 67(8), pp. 7314-7325, 2018.
\bibitem{b11} R. Ayyalasomayajula, A. Arun, C. Wu, S. Sharma, A. Sethi, D. Vasisht, and D. Bharadia, ``Deep learning based wireless localization for indoor navigation,'' Proceedings of the 26th Annual International Conference on Mobile Computing and Networking, pp. 1-14, 2020.
\bibitem{b12} J. Yim, S. Jeong, K. Gwon, and J. Joo, ``Improvement of Kalman filters for WLAN based indoor tracking,'' Expert Systems with Applications vol. 37(1), pp. 426-433, 2010.
\bibitem{b13} J. Yim, ``Introducing a decision tree-based indoor positioning technique. Expert Systems with Applications,'' Computer Science vol. 34(2), pp. 1296-1302, 2008.
\bibitem{b14} P. Sthapit, H. Gang, and J. Pyun, ``Bluetooth based indoor positioning using machine learning algorithms,'' 2018 IEEE International Conference on Consumer Electronics-Asia (ICCE-Asia), pp. 206-212, 2018.
\bibitem{b15} A. Koutris, T. Siozos, Y. Kopsinis, A. Pikrakis, T. Merk, M. Mahlig, ... and P. Karlsson, ``Deep Learning-Based Indoor Localization Using Multi-View BLE Signal,'' Sensors, vol. 22(7), pp. 2759, 2022.
\bibitem{b16} I. Cox, ``Blanche: Position estimation for an autonomous robot vehicle,'' Autonomous robot vehicles, pp. 221-228, 1990.
\bibitem{b17} T. Umeno and E. Hayashi, ``Development of an autonomous personal robot: the visual processing system for autonomous driving,'' Proceedings of the 10th International Symposium on Artificial Life and Robotics (AROB10), Beppu, Oita, Japan, p. 4, 2005.
\bibitem{b18} E. Hayashi, ``Navigation system for an autonomous robot using an ocellus camera in an indoor environment,'' Artificial Life and Robotics, vol. 12(1), pp. 346-352, 2008.
\bibitem{b19} A. Abdalwhab and H. Liu, ``Zero-shot object detection for indoor robots,'' 2019 International Joint Conference on Neural Networks (IJCNN) , pp. 1-8, 2019.
\bibitem{b20} T. Khanh, T. Hai, V. Nguyen, T. Nguyen, N. Thu, and E. Huh, ``The Practice of Cloud-based Navigation System for Indoor Robot,'' 2020 14th International Conference on Ubiquitous Information Management and Communication (IMCOM), pp. 1-4, 2020.
\bibitem{b21} N. Swift, ``Easy A* (star) Pathfinding,'' Medium, \url{https://medium.com/@nicholas.w.swift/easy-a-star-pathfinding-7e6689c7f7b2}. Accessed December 2022.
\bibitem{b22} S. Tiku, D. Gufran, S. Pasricha, “Multi-Head Attention Neural Network for Smartphone Invariant Indoor Localization”, IEEE Conference on Indoor Positioning and Indoor Navigation (IPIN), 2022

\end{thebibliography}
\end{document}